\documentclass[runningheads]{llncs}
\usepackage[T1]{fontenc}
\usepackage{graphicx,verbatim}
\usepackage{amssymb}
\usepackage{multirow}
\usepackage{subcaption}
\usepackage{amsmath}
\usepackage{xcolor}
\usepackage{booktabs}
\usepackage{pgfplots}
\pgfplotsset{compat=1.18}
\usepackage{pgfplotstable}
\usepgfplotslibrary{groupplots}

\definecolor{benign}{RGB}{0,160,0}
\definecolor{malignant}{RGB}{180,0,0}

\begin{document}
\title{Geometry-Aware Superpixel Graph Transformer with Metadata for Skin Lesion Classification}

\titlerunning{GeoMeta-GT: Geometry-Aware Graph Transformer}

\author{Muhammad Azeem\orcidID{0009-0003-7713-6926} \and
Tanveer Hussain\orcidID{0000-0003-4861-8347} \and
Amr Ahmed\orcidID{0000-0002-7749-7911} \and
Ardhendu Behera\thanks{Corresponding author (beheraa@edgehill.ac.uk).}\orcidID{0000-0003-0276-9000}}

\authorrunning{M. Azeem et al.}

\institute{Edge Hill University, Ormskirk, Lancashire, L39 4QP, United Kingdom\\
\email{\{shoukatm, hussaint, ahmeda, beheraa\}@edgehill.ac.uk}}

\maketitle              
\begin{abstract}
Automated skin cancer classification from dermoscopic images remains challenging due to heterogeneous lesion structure, strong intra-class variability, and subtle visual differences between benign and malignant cases. Existing CNN/ViT pipelines typically rely on global or patch-level features and often combine patient metadata via late fusion, which limits spatially grounded multimodal reasoning. We present a novel \textbf{region-based graph learning framework} that explicitly models lesions as \textbf{graphs of spatially coherent superpixel regions} represented as \emph{frozen CNN features}. To capture fine-grained lesion arrangements, we encode inter-regional geometry as edge attributes and introduce a dedicated \textbf{metadata context node} connected to all regions, providing structured integration of demographic/clinical variables within the same relational space. Node representations are updated using our \emph{edge-aware graph transformer} followed by \emph{attention-driven propagation}, and a final graph-level embedding for benign–malignant classification. Experiments on four public benchmarks demonstrate that explicit region-level relational modeling and graph-native multimodal fusion yield consistent gains over the state-of-the-art. Consequently, we establish a new graph-centric perspective in which CNN features are modeled as relational nodes and improved through contextual integration, yielding more expressive and robust classifications.

\keywords{Skin Lesion Classification \and Superpixel Graph Transformer \and Spatial Relational Modeling \and Clinical Metadata Integration.}
\end{abstract}
\section{Introduction}
Skin cancer is among the most prevalent malignancies, and melanoma remains the most lethal subtype, making an early and reliable diagnosis critical to patient outcomes \cite{siegel2026cancer}. Deep learning has substantially improved dermoscopic lesion classification, supported by large-scale datasets such as ISIC \cite{kurtansky2024slice} and HAM10000 \cite{tschandl2018ham10000}. However, dermoscopic lesions are intrinsically heterogeneous; clinically meaningful cues often emerge from the co-occurrence and spatial arrangement of multiple subregions (e.g., pigment network, globules, streaks, regression structures, and boundary irregularity). Many CNN and ViT pipelines still represent an image as a global vector or as largely independent grid/patch tokens, which can weaken their ability to explicitly model region-to-region interactions and the geometry of subtle variations \cite{azeem2023neural,naseri2025diagnosis,LMSViT2025}. In addition, patient metadata (e.g., age, sex, anatomical site, clinical context) is a known source of complementary information but is frequently fused late (e.g., concatenated in the classifier), limiting its ability to condition where and how visual evidence is aggregated \cite{ahammed2025cvpr}.

Recent work has explored diverse architectures for dermoscopy, including lightweight and hybrid CNNs variants \cite{tuncer2024lightweight,pradhan2024skin,tahir2023dscc_net,azeem2023skinlesnet}, attention-enhanced CNNs and sequential hybrids \cite{datta2021soft,abohashish2025enhanced,li2024self}, and transformer-based models designed to improve global context modeling \cite{xin2022improved}. Multimodal approaches have also gained traction, ranging from contrastive and dual-stage fusion to personalized multimodal learning \cite{dai2023deeply,christopoulos2025skin,fan2025personalized}. While these methods demonstrate progress, two limitations remain prominent: (i) most models operate on grid/patch tokens and do not provide an explicit, spatially grounded representation of lesion subregions and their geometric relations; and (ii) metadata are commonly treated as an auxiliary vector rather than being integrated into the core reasoning process, reducing interpretability and limiting structured multimodal interactions. Graph neural networks (GNNs) offer a natural abstraction for region-centric reasoning by encoding regions in images as nodes and edges \cite{kipf2016semi}. However, only a few dermoscopy graphs employ context-aware graph reasoning \cite{azeem2026context}, leaving a gap for \emph{edge-attributed} \emph{geometry-aware attention} with \emph{metadata integrated in the graph}.

To address this gap, we propose \textbf{GeoMeta-GT}, a Geometry-Aware Superpixel Graph Transformer with Metadata for skin lesion classification. Each dermoscopic image is decomposed into superpixels to form region nodes, with node descriptors extracted from a frozen CNN.
Fine-grained lesion organization is encoded through geometric edge attributes derived from inter-regional distance and orientation. Patient metadata is embedded as a dedicated context node connected to all regions, enabling message-passing-based fusion rather than late concatenation. A novel edge-aware graph transformer performs locally adaptive aggregation along geometry-informed edges, followed by attentional refinement and graph-level pooling for the final prediction.

\noindent \textbf{Contributions.} (1) A parameter-efficient, spatially grounded and multimodally structured approach for robust dermoscopic lesion classification; (2) a geometry-attributed superpixel graph representation with metadata-as-node fusion for dermoscopy; (3) an edge-aware graph transformer for spatially grounded, diagnostically focused region aggregation; (4) a similarity-weighted refinement module to harmonize semantically consistent regions; and (5) consistent improvements across multiple public benchmarks using a frozen backbone.

\section{Methodology}
\begin{figure}
\includegraphics[width=\textwidth]{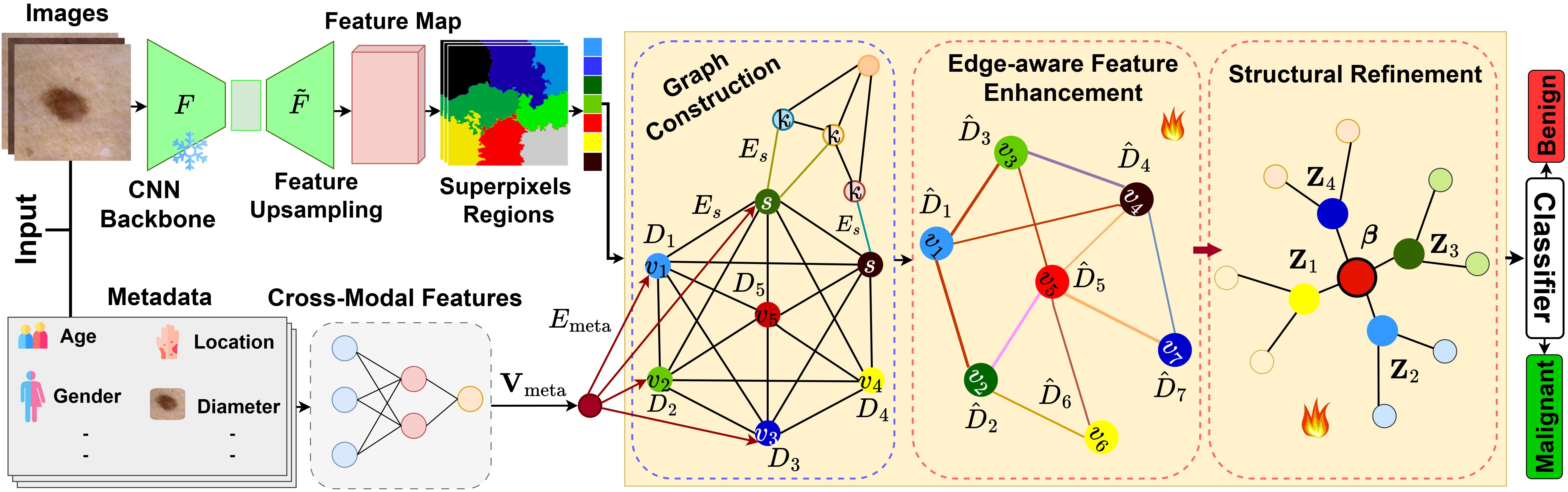}
\caption{\textbf{Overview of GeoMeta-GT}. The input dermoscopic image is encoded by a frozen CNN and partitioned into SLIC superpixels to form region nodes $V_S$ with pooled deep descriptors $D_S$. Region-to-region edges $E_S$ are enriched with geometric attributes (distance and orientation), while patient metadata is represented as a context node $V_\text{meta}$ connected to all regions (nodes $V_S$). An edge-aware graph transformer and similarity-weighted structural refinement perform relational reasoning, followed by global pooling for skin lesion classification.}
\label{model_fig}
\end{figure}

Given a dermoscopic image and associated patient metadata, the \textbf{GeoMeta-GT} pipeline (Fig. \ref{model_fig}) constructs a \textit{geometry-attributed superpixel graph} and performs \textit{edge-aware attention-based message passing} to classify lesions as benign or malignant. The CNN backbone remains frozen, and all task-specific adaptation is confined to the graph module. 

\noindent\textbf{Superpixel decomposition and region descriptors.} Let $I \in \mathbb{R}^{H \times W \times 3}$ denote an input image. 
We first extract deep feature representations using a pretrained frozen CNN, $F = f_{\text{CNN}}(I) \in \mathbb{R}^{C \times h \times w}$, where $C$ is the number of channels and $h \times w$ is the spatial resolution of the feature map. Since $h < H$ and $w < W$, we upsample $F$ to the original image size using bilinear interpolation, $\tilde{F} = \text{Interp}(F) \in \mathbb{R}^{C \times H \times W}$. In dermoscopic skin lesion imaging, the raw RGB intensities are often poorly aligned with the underlying anatomy or pathology; in particular, Intensity similarity $\not\Rightarrow$ anatomical or pathological similarity. Instead, pretrained CNN defines a task-dependent embedding space in which spatial locations that share similar anatomical or pathological characteristics are mapped to nearby feature vectors (feature similarity $\approx$ anatomical, pathological, and task-relevant semantic proximity). Consequently, performing clustering in this deep feature space produces image segments that align more closely with the true lesion structure than those obtained by clustering directly in the raw intensity domain. We therefore apply simple linear iterative clustering (SLIC) \cite{achanta2012slic} directly on the features $\tilde{F}$ to decompose the image into $K$ superpixels, $R = \{R_1, R_2, \dots, R_K\}$, where each superpixel $R_k$ is a spatially connected cluster whose pixels exhibit similar deep feature vectors under SLIC's joint feature--spatial distance metric. To obtain superpixel-level descriptors, we aggregate the deep features within each region. Specifically, for each superpixel $R_k$, we form a compact descriptor by concatenating a global context statistic (mean-pooled) and a local saliency statistic (max-pooled) features over all pixels in $R_k$:

\begin{equation}
\label{eqn:feat_space}
D_k =
\underbrace{
\frac{1}{|R_k|} \sum_{(i,j)\in R_k} \tilde{F}(:, i, j)
}_{\text{Mean pooling}}
\;\Bigg\|\;
\underbrace{
\max_{(i,j)\in R_k} \tilde{F}(:, i, j)
}_{\text{Max pooling}}
\in \mathbb{R}^{2C},
\end{equation}

\noindent where $\|$ denotes channel-wise concatenation. The resulting descriptor $D_k$ is used as the node feature of the $k^{\text{th}}$ superpixel. This parameter-free aggregation provides stable, computationally efficient region representations, which are directly used as node features in the subsequent superpixel graph transformer.

\noindent\textbf{Graph construction with geometric and metadata nodes.} We construct a graph $G = (V, E)$ consisting of a set of superpixel nodes $V_S = \{v_1, \dots, v_K\}$, each associated with a corresponding node feature $D_S = \{D_1, \dots, D_K\}$, and a single metadata node $V_{\text{meta}}$. Each superpixel node $v_k\in V_S$ is associated with a 2D image coordinate $p_k = (x_k, y_k) \in \mathbb{R}^2$, defined as the center of the axis-aligned bounding box of the corresponding SLIC region detected in the previous step.

\noindent The spatial edges between the superpixel nodes, denoted by $E_S$, are constructed by $n$-nearest neighbor search. For each superpixel node $v_k \in V_S$ with centroid position $p_k$, we identify its $n$ closest superpixel neighbors $v_s \in V_S,\, s \neq k$, based on the Euclidean distance between their centroid locations $p_s$ and $p_k$. Repeating this procedure for all $v_k \in V_S$ yields the set of directed superpixel-to-superpixel edges $E_S$. For each edge $e_{k,s} \in E_S$ connecting node $k$ to node $s$, we define a geometric edge attribute consisting of: 1) spatial distance $d_{k,s} = \lVert p_s - p_k \rVert_2$, and 2) relative orientation $\theta_{k,s} = \operatorname{atan2}(y_s - y_k, x_s - x_k)$.

\noindent\textbf{Metadata-as-node fusion.} Let \(\mathbf{m} \in \mathbb{R}^{d}\) denote the raw patient-level metadata associated with an image. We first standardize all numeric fields via \(z\)-score normalization and one-hot encode all categorical fields, producing a concatenated metadata vector \(\hat{\mathbf{m}} \in \mathbb{R}^{d_m}\). We then embed these metadata into the superpixel node feature space \(D_k\) (cf. Eq.~(\ref{eqn:feat_space})) using a learnable linear projection $D_{\text{meta}} \rightarrow \phi(\hat{\mathbf{m}}) = W_m \hat{\mathbf{m}} + b_m \in \mathbb{R}^{2C}$, where \(W_m\) and \(b_m\) are trainable parameters to match the dimensionality of the superpixel node features (\(2C\)).

\noindent We introduce a dedicated metadata node \(V_{\text{meta}}\) with feature vector \(D_{\text{meta}}\) and connect it to all superpixel nodes \(V_S\). Formally, we define the metadata edges as \(E_{\text{meta}} = \{(V_{\text{meta}}, k), (k, V_{\text{meta}}) \mid k \in V_S\}\).
The final graph $G$ is thus given by nodes \(V = V_S \cup V_{\text{meta}}\), with the respective feature \(D = D_S \cup D_{\text{meta}}\) and edge set \(E = E_S \cup E_{\text{meta}}\). This serves as input to the downstream graph neural network. The GNN performs relational reasoning on this graph $G$ to jointly update all node features, thereby integrating image-derived superpixel information with patient-level metadata in a unified representation.

\subsection{Edge-aware Graph Transformer for Feature Enhancement}
We propose a novel \emph{edge-aware graph transformer} to perform relational reasoning by explicitly incorporating edge attributes into the attention mechanism. Our method advances the local adaptive feature representation~\cite{shi2020masked} in the graph domain, improving the $i^{\text{th}}$ superpixel node feature descriptor $D_i \in D$ by selectively aggregating information from its adjacent superpixels.

\noindent For node $i$, we define a single attention head (multi-head omitted for clarity) as follows. Firstly, node features $(D_i, D_j)$ and edge attributes $(e_{i,j})$ are linearly projected using learnable weight matrices $W_1, W_2$, and $W_3$: query $
\mathcal{Q}_i = W_1 D_i, 
\mathcal{K}_j = W_2 D_j, 
\hat{e}_{i,j} = W_3 e_{i,j}$,
where $\mathcal{Q}_i$ and $\mathcal{K}_j$ denote the query and key vectors, and $\hat{e}_{i,j}$ is the embedded edge attribute.

\noindent The edge-aware attention score $\alpha_{i,j}$ from node $i$ (target) to node $j$ (neighbor) is then calculated by injecting the edge embedding into the key before the dot-product:

\begin{equation}
\alpha_{i,j} = \operatorname{softmax}\left( \frac{\mathcal{Q}_i \bigodot  (\mathcal{K}_j + \hat{e}_{i,j})}{\sqrt{d}} \right),
\end{equation}
where $d$ is the hidden dimension per head and $\bigodot$ denotes the dot product. 
Next, we update the feature at the node $i$ by aggregating messages from its neighbors. To preserve the semantic content of the edge in the output representation, we re-inject the edge embedding into the message:

\begin{equation}
\label{eqn:embed}
\hat{D}_{i} = \mathcal{Q}_i + \sum_{j \in \mathcal{N}(i)} \alpha_{i,j} \bigl(W_4 D_j + W_5 e_{i,j} \bigr),
\end{equation}

\noindent where $\mathcal{N}(i)$ denotes the set of neighbors of node $i$, and $W_4$ and $W_5$ are learnable weight matrices for node and edge messages, respectively.

\noindent This design offers two key advantages. (1) Since $e_{i,j}$ participates directly in the query--key interaction, the model learns to emphasize edges that are most informative for the downstream task, rather than treating all connections uniformly. (2) The edge attributes can effectively ``shut down'' or ``boost'' the influence of specific neighbors, inducing locally adaptive region features that focus on diagnostically relevant interactions (e.g., irregular borders and varied pigmentation), while suppressing contributions from noisy or less informative neighbors.

\subsection{Structural Refinement and Graph-level Readout}
To encourage semantic consistency among connected superpixels, we introduce a lightweight similarity-weighted propagation step applied to $\hat{D}_i$ in Eq. (\ref{eqn:embed}). This module refines node embeddings by aggregating information from neighboring nodes based on feature similarity, thereby ``denoising'' the representations produced by the edge-aware graph transformer while limiting the risk of overfitting. Concretely, we employ a learnable temperature parameter $\beta$ that controls the sharpness of the similarity distribution and thus how selectively each node aggregates information from its neighborhood. Our design is inspired by the similarity-based propagation mechanism in \cite{thekumparampil2018attention}, which allows visually or semantically similar superpixels to exchange information even when they are spatially distant. This is particularly beneficial in dermoscopic images, where malignant structures may appear in multiple disjoint regions of a lesion.

\noindent
Given the $i^{\text{th}}$ node embedding $\hat{D}_i$, we compute the refined embedding $Z_i$ as
\begin{equation}
Z_i = \sum_{j\in\mathcal{N}(i)\cup\{i\}} \gamma_{i,j} \, \hat{D}_j, \quad
\gamma_{i,j} = \operatorname{softmax}_{j\in\mathcal{N}(i)\cup\{i\}} \bigl( \beta \, cosine(\hat{D}_i, \hat{D}_j) \bigr).
\end{equation}
This refinement step improves the discriminative patterns that recur in the lesion, while suppressing inconsistent or spurious activations. 
Finally, we obtain a graph-level representation via global mean pooling over the refined node embeddings, $Z = \operatorname{MeanPool}(\{Z_i\}_{i \in V})$,
which is subsequently fed into a classifier for binary lesion classification.

\section{Experiments and Results}
\textbf{Training Details.} The model is implemented in the PyTorch framework and trained on a single NVIDIA GPU using the Adam optimizer with an initial learning rate of $1\times10^{-3}$, a weight decay of $1\times10^{-4}$, and a ReduceLROnPlateau scheduler for 50 epochs. The CNN backbone remains frozen, and only the graph transformer and classifier parameters are optimized, ensuring efficient training and fair, consistent evaluation across datasets. 

\noindent \textbf{Datasets.} We evaluated the proposed method on four public dermoscopic benchmarks, \textbf{ISIC2024} \cite{kurtansky2024slice}, \textbf{HAM10000} \cite{tschandl2018ham10000}, \textbf{PAD-UFES-20} \cite{pacheco2020pad}, and \textbf{HIBA} \cite{ricci2023dataset}, covering diverse imaging conditions and lesion distributions. Following state-of-the-art (SOTA) practices, all images are resized to $224 \times 224$, and the dataset is split into 80\% for training and 20\% for testing to enable a fair comparison.

\noindent\textbf{Quantitative Evaluation and Discussion.}  
The proposed \textbf{GeoMeta-GT} consistently outperformed SOTA methods (Table \ref{full_sota}) on four benchmarks, demonstrating both robustness and generalization. On the ISIC2024 dataset, GeoMeta-GT achieves 98.61\% accuracy, substantially exceeding the best competing multimodal contrastive model (96.69\%) \cite{christopoulos2025skin}. This gain indicates that explicitly modeling spatial relationships and integrating metadata as a dedicated graph node yields more discriminative and context-aware representations than contrastive image with metadata embedding alone. On HAM10000, the model's 98.23\% surpassed the strong hybrid CNN–LSTM model (96.21\%) \cite{abohashish2025enhanced} and transformer-based approaches (e.g., 94.31\%) \cite{xin2022improved}, highlighting that structured superpixel graphs with edge-aware attention capture fine-grained lesion structure and inter-region morphology better than sequential or global attention without explicit connectivity. On PAD-UFES-20, GeoMeta-GT’s 97.17\% accuracy exceeds multimodal learning baselines (95.61\%) \cite{fan2025personalized} and context-aware GNNs (94.29\%) \cite{azeem2026context}, showing stronger robustness to variability and heterogeneity in lesion appearances and imaging conditions. The consistently higher recall and F1-scores further reflect improved sensitivity to diverse lesion features. On the more challenging HIBA dataset, GeoMeta-GT’s 95.41\% performance markedly outperformed recent graph-based and deep learning models such as context-aware GNNs (89.19\%) \cite{azeem2026context} and multimodal contrastive learning (87.51\%) \cite{christopoulos2025skin}, demonstrating superior ability to handle inconsistent acquisition and imaging noise. Across all benchmarks, the empirical improvements confirm that combining geometric edge encoding, structured metadata-as-node fusion, and edge-aware relational attention produced more expressive and generalizable representations than pure CNN, transformer, sequential, or shallow GNN alternatives. These gains are especially notable on heterogeneous and challenging datasets (PAD-UFES-20 and HIBA), underpinning the practical value of geometry-aware relational reasoning in clinical skin lesion classification.

\begin{table}[t]
\centering
\caption{Comparison (\%) with SOTA methods across four benchmark datasets. Best are shown in \textbf{bold.}}
\label{full_sota}
\setlength{\tabcolsep}{3pt}
\resizebox{\textwidth}{!}{
\begin{tabular}{l l l c c c c}
\hline
Dataset & Method & Accuracy & Precision & Recall & F1-score \\
\hline

\multirow{8}{*}{ISIC2024 \cite{kurtansky2024slice}}
 & Deep CNN \cite{golnoori2023metaheuristic} & 90.10 & 89.81 & 90.10 & 89.81 \\
 & Lightweight CNN \cite{tuncer2024lightweight} & 92.12 & 92.10 & 92.11 & 92.01 \\
 & Deep Framework \cite{ozdemir2025robust} & 93.48 & 93.24 & 93.41 & 92.72 \\
 & Ensemble Model \cite{imran2022skin} & 93.51 & 93.31 & 93.51 & 92.81 \\
 & DSCC\_Net \cite{tahir2023dscc_net} & 94.17 & 94.17 & 93.76 & 93.93 \\
 & Hybrid CNN \cite{pradhan2024skin} & 94.44 & 94.38 & 94.21 & 94.44 \\
 & Multimodal Contrastive \cite{christopoulos2025skin} & 96.69 & 96.14 & 96.41 & 96.69 \\
 & \textbf{GeoMeta-GT} & \textbf{98.61} & \textbf{98.58} & \textbf{98.64} & \textbf{98.61} \\
\hline

\multirow{8}{*}{HAM10000 \cite{tschandl2018ham10000}}
 & Context-aware GNN \cite{azeem2026context} & 86.69 & 90.01 & 89.19 & 89.19 \\
 & MultiExCam \cite{ruga2025multiexcam} & 87.54 & 87.45 & 87.49 & 87.54 \\
 & Fused Architecture \cite{arshad2025multiclass} & 91.31 & 91.21 & 91.45 & 91.30 \\
 & Soft-Attention Model \cite{datta2021soft} & 93.71 & 93.70 & 93.01 & 93.21 \\
 & Transformer Network \cite{xin2022improved} & 94.31 & 94.10 & 95.01 & 93.20 \\
 & Inception-ResNet \cite{bozkurt2023skin} & 95.09 & 95.49 & 95.16 & 95.27 \\
 & Hybrid LSTM--CNN \cite{abohashish2025enhanced} & 96.21 & 94.75 & 94.93 & 95.55 \\
 & \textbf{GeoMeta-GT} & \textbf{98.23} & \textbf{98.21} & \textbf{98.26} & \textbf{98.23} \\
\hline

\multirow{8}{*}{PAD-UFES-20 \cite{pacheco2020pad}}
 & MDFNet \cite{chen2023mdfnet} & 80.42 & 78.02 & 79.97 & 78.64 \\
 & Multimodal Dual-Stage \cite{dai2023deeply} & 85.11 & 84.31 & 84.61 & 85.11 \\
 & CNN Model \cite{li2024self} & 86.41 & 90.01 & 85.30 & 85.91 \\
 & CNN-Attention Hybrid \cite{pacheco2021attention} & 91.22 & 91.35 & 91.41 & 91.19 \\
 & Multimodal Contrastive \cite{christopoulos2025skin} & 93.51 & 93.41 & 93.21 & 93.25 \\
 & Context-aware GNN \cite{azeem2026context} & 94.29 & 94.40 & 94.29 & 94.26  \\
 & Multimodal Learning \cite{fan2025personalized} & 95.61 & 95.54 & 95.12 & 95.61 \\
 & \textbf{GeoMeta-GT} & \textbf{97.17} & \textbf{97.17} & \textbf{97.21} & \textbf{97.17} \\
\hline

\multirow{7}{*}{HIBA \cite{ricci2023dataset}}
 & Diffusion Models \cite{uliana2025diffusion} & 70.68 & 78.94 & 62.36 & 62.59 \\
 & MM-Skin \cite{araujo2024key} & 81.65 & 81.41 & 81.65 & 81.58 \\
 & CNN Models \cite{oyedeji2024clinical} & 82.71 & 83.21 & 71.11 & 81.71 \\
 & MultiExCam \cite{ruga2025multiexcam} & 85.11 & 86.20 & 84.45 & 85.11 \\
 & Multimodal Contrastive \cite{christopoulos2025skin} & 87.51 & 86.40 & 87.51 & 88.51 \\
 & Context-aware GNN \cite{azeem2026context} & 89.19 & 90.01 & 89.19 & 89.19 \\
 & \textbf{GeoMeta-GT} & \textbf{95.41} & \textbf{95.38} & \textbf{95.44} & \textbf{95.41} \\
\hline

\end{tabular}
} 
\end{table}

\section{Ablation Studies}
Comprehensive ablations validate the contribution of each component. The results consistently show that geometric encoding, metadata fusion, backbone depth, and attention design collectively drive the proposed model’s strong and stable performance across all datasets.

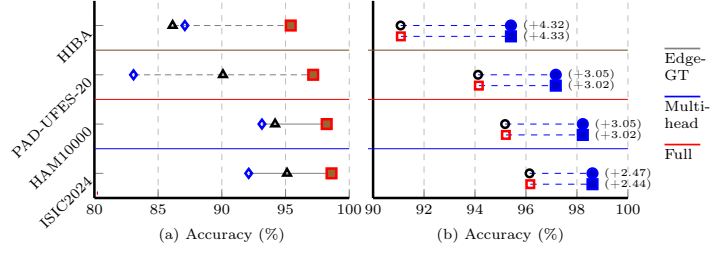
\begin{figure}[t]
\centering
\resizebox{0.8\linewidth}{!}{
\pgfplotsset{compat=1.17}

\begin{tikzpicture}

\begin{axis}[
    name=axa,
    xmin=80, xmax=100,
    width=0.48\linewidth,
    height=0.40\linewidth,
    axis x line*=bottom,
    axis y line*=left,
    axis line style={very thick},
    xmajorgrids,
    grid style={dashed},
    xlabel={(a) Accuracy (\%)},
    ymin=0.5, ymax=4.5,
    ytick={1,2,3,4},
    yticklabels={ISIC2024,HAM10000,PAD-UFES-20,HIBA},
    yticklabel style={font=\scriptsize, rotate=45, anchor=east},
    label style={font=\scriptsize},
    tick label style={font=\scriptsize},
    clip=false
]

\addplot+[no marks, very thin] coordinates {(80,1.5) (100,1.5)};
\addplot+[no marks, very thin] coordinates {(80,2.5) (100,2.5)};
\addplot+[no marks, very thin] coordinates {(80,3.5) (100,3.5)};

\addplot+[no marks, thin, gray] coordinates {(92.11,1) (98.61,1)};
\addplot+[no marks, thin, gray] coordinates {(93.15,2) (98.23,2)};
\addplot+[no marks, thin, gray] coordinates {(83.07,3) (97.17,3)};
\addplot+[no marks, thin, gray] coordinates {(86.14,4) (95.41,4)};

\addplot+[black, very thick, mark=none] coordinates {(80.2,0.6) (80.2,0.6)};
\addplot+[blue, very thick, mark=none]  coordinates {(80.2,0.6) (80.2,0.6)};
\addplot+[red, very thick, mark=none]   coordinates {(80.2,0.6) (80.2,0.6)};

\addplot+[only marks, mark=triangle, mark size=2.1pt, very thick, black] coordinates {
(95.12,1) (94.18,2) (90.09,3) (86.14,4)
};
\addplot+[only marks, mark=diamond, mark size=2.1pt, very thick, blue] coordinates {
(92.11,1) (93.15,2) (83.07,3) (87.10,4)
};
\addplot+[only marks, mark=square*, mark size=2.3pt, very thick, red] coordinates {
(98.61,1) (98.23,2) (97.17,3) (95.41,4)
};

\end{axis}

\begin{axis}[
    name=axb,
    at={(axa.east)},
    anchor=west,
    xshift=4mm,
    xmin=90, xmax=100,
    width=0.48\linewidth,
    height=0.40\linewidth,
    axis x line*=bottom,
    axis y line*=left,
    axis line style={very thick},
    xmajorgrids,
    grid style={dashed},
    xlabel={(b) Accuracy (\%)},
    ymin=0.5, ymax=4.5,
    ytick={1,2,3,4},
    yticklabels={,,,},
    ytick style={draw=none},
    yticklabel style={font=\scriptsize},
    label style={font=\scriptsize},
    tick label style={font=\scriptsize},
    clip=false
]

\addplot+[no marks, very thin] coordinates {(89.8,1.5) (100,1.5)};
\addplot+[no marks, very thin] coordinates {(89.8,2.5) (100,2.5)};
\addplot+[no marks, very thin] coordinates {(89.8,3.5) (100,3.5)};

\addplot+[only marks, mark=o, mark size=1.8pt, very thick, black] coordinates {
(96.14,1) (95.18,2) (94.12,3) (91.08,4)
};
\addplot+[only marks, mark=*, mark size=2.3pt, very thick, blue] coordinates {
(98.61,1) (98.23,2) (97.17,3) (95.41,4)
};
\addplot+[no marks, thin, blue, dashed] coordinates {(96.14,1) (98.61,1)};
\addplot+[no marks, thin, blue, dashed] coordinates {(95.18,2) (98.23,2)};
\addplot+[no marks, thin, blue, dashed] coordinates {(94.12,3) (97.17,3)};
\addplot+[no marks, thin, blue, dashed] coordinates {(91.08,4) (95.41,4)};

\addplot+[only marks, mark=square, mark size=1.8pt, very thick, red] coordinates {
(96.17,0.78) (95.21,1.78) (94.15,2.78) (91.09,3.78)
};
\addplot+[only marks, mark=square*, mark size=2.3pt, very thick, blue] coordinates {
(98.61,0.78) (98.23,1.78) (97.17,2.78) (95.41,3.78)
};
\addplot+[no marks, thin, blue, dashed] coordinates {(96.17,0.78) (98.61,0.78)};
\addplot+[no marks, thin, blue, dashed] coordinates {(95.21,1.78) (98.23,1.78)};
\addplot+[no marks, thin, blue, dashed] coordinates {(94.15,2.78) (97.17,2.78)};
\addplot+[no marks, thin, blue, dashed] coordinates {(91.09,3.78) (95.41,3.78)};

\node[anchor=west, font=\tiny] at (axis cs:98.61,1) {~(+2.47)};
\node[anchor=west, font=\tiny] at (axis cs:98.23,2) {~(+3.05)};
\node[anchor=west, font=\tiny] at (axis cs:97.17,3) {~(+3.05)};
\node[anchor=west, font=\tiny] at (axis cs:95.41,4) {~(+4.32)};
\node[anchor=west, font=\tiny] at (axis cs:98.61,0.78) {~(+2.44)};
\node[anchor=west, font=\tiny] at (axis cs:98.23,1.78) {~(+3.02)};
\node[anchor=west, font=\tiny] at (axis cs:97.17,2.78) {~(+3.02)};
\node[anchor=west, font=\tiny] at (axis cs:95.41,3.78) {~(+4.33)};

\end{axis}

\node[
    anchor=south,
    font=\scriptsize
] at ([xshift=10mm, yshift=-12mm]axb.east) {
\begin{tabular}{l}
\textcolor{gray}{\rule{6mm}{0.8pt}}\\[-0.5mm]
Edge-\\GT \\[0.1mm]
\textcolor{blue}{\rule{6mm}{0.8pt}}\\[-0.5mm]
Multi-\\
head \\[0.1mm]
\textcolor{red}{\rule{6mm}{0.8pt}}\\[-0.5mm]
Full
\end{tabular}
};

\end{tikzpicture}
}

\caption{Illustrates performance comparison and ablation analysis. (a) Accuracy only of model variants. (b) Impact of GEE (dumbbells) on Accuracy and F1-score, showing consistent and substantial gains across all datasets.}
\label{ab_combined_final}
\end{figure}

\noindent\textbf{Impact of Geometric Edge Encoding.} Incorporating geometric edge encoding (GEE) leads to consistent and substantial improvements across all datasets, as shown in Fig. \ref{ab_combined_final}(b). Accuracy increases from 96.14\% to 98.61\% on ISIC2024 (+2.47\%), with the F1 score from 96.17\% to 98.61\% (+2.44\%). On HAM10000, accuracy rises from 95.18\% to 98.23\% (+3.05\%), and the F1 score from 95.21\% to 98.23\% (+3.02\%). On PAD-UFES-20, accuracy improves from 94.12\% to 97.17\% (+3.05\%), with the F1 score increasing from 94.15\% to 97.17\% (+3.02\%). On HIBA, accuracy advances from 91.08\% to 95.41\% (+4.33\%), while the F1 score rises from 91.09\% to 95.41\% (+4.32\%). These stable margins across datasets confirm that modeling spatial distance and relative orientation significantly enhances relational expressiveness, making GEE a key contributor to the model’s superior performance and generalization.

\noindent\textbf{Model Variations.}  
Using only the edge-aware graph transformer yields 95.12\%, 94.18\%, 90.09\%, and 86.14\% accuracy on ISIC2024, HAM10000, PAD-UFES-20, and HIBA, while naive multi-head concatenation further drops performance to 92.11\%, 93.15\%, 83.07\%, and 87.10\%. In contrast, the full model achieves 98.61\%, 98.23\%, 97.17\%, and 95.41\%, confirming that integrating edge-aware attention, structural refinement, and metadata fusion is crucial for optimal and stable performance, shown in Fig. \ref{ab_combined_final}(a).

\noindent\textbf{Significance of Metadata.}  
Incorporating patient metadata substantially improves accuracy from 94.12\%, 95.16\%, 92.08\%, and 90.14\% to 98.61\%, 98.23\%, 97.17\%, and 95.41\% on ISIC2024, HAM10000, PAD-UFES-20, and HIBA, respectively. These consistent gains confirm that structured metadata fusion significantly enhances discriminative power and generalization.

\noindent\textbf{Evaluating Backbone Sensitivity.}  
Backbone selection strongly affects performance, with VGG19 achieving 91.08\%, 92.12\%, 89.09\%, and 87.06\%, Inception reaching 93.12\%, 94.14\%, 91.10\%, and 88.07\%, and EfficientNet improving to 95.14\%, 96.16\%, 94.12\%, and 92.09\% on ISIC2024, HAM10000, PAD-UFES-20, and HIBA, respectively. ResNet152 further elevates accuracy to 98.61\%, 98.23\%, 97.17\%, and 95.41\%, confirming that deeper residual features provide the most discriminative region representations for the proposed graph framework.

\section{Conclusion}
We presented a geometry-aware superpixel graph framework for skin lesion classification that models dermoscopic images as region adjacency graphs with edge-aware relational reasoning. Using frozen CNN features and integrating patient metadata as a dedicated graph node, the model captures both structural and clinical context without costly backbone fine-tuning. Extensive experiments and ablations confirm its robustness and superior generalization across benchmarks. Future work will extend the approach to multi-class diagnosis, uncertainty modeling, and enhanced interpretability for clinical deployment. Code is available at: \url{https://github.com/azeemchaudharyg/GeoMeta-GT}


\bibliographystyle{splncs04}
\bibliography{refs}
\end{document}